\def\paperID{0000}
\def\confName{ICCV}
\def\confYear{2025}

\def\paperTitle{Running VLAs at Real-time Speed}

\def\authorBlock{
    Yunchao Ma$^1$ \quad
    Yizhuang Zhou$^2$ \quad
    Yunhuan Yang$^1$ \quad
    Tiancai Wang$^1$ \quad
    Haoqiang Fan$^1$ \\
    $^1$ Dexmal \qquad $^2$ StepFun \\
    {\tt\small \{myc, yyh, wtc, fhq\}@dexmal.com jupiter@stepfun.com}
}

\newif\ifreview 
\newif\ifarxiv \newcommand{\arxiv}{\arxivtrue}
\newif\ifcamera 
\newif\ifrebuttal 

\arxiv 

\pdfoutput=1
\documentclass[10pt,twocolumn,letterpaper]{article}
\ifreview \usepackage[review]{cvpr} \fi
\ifarxiv \usepackage[pagenumbers]{cvpr} \fi
\ifrebuttal \usepackage[rebuttal]{cvpr} \fi
\ifcamera \usepackage{cvpr} \fi

\usepackage{graphicx}	
\usepackage{amsmath}	
\usepackage{amssymb}	
\usepackage{booktabs}
\usepackage{times}
\usepackage{microtype}
\usepackage{epsfig}
\usepackage{caption}
\usepackage{float}
\usepackage{placeins}
\usepackage{color, colortbl}
\usepackage{stfloats}
\usepackage{enumitem}
\usepackage{tabularx}
\usepackage{xstring}
\usepackage{multirow}
\usepackage{xspace}
\usepackage{url}
\usepackage{subcaption}
\usepackage{xcolor}
\usepackage[hang,flushmargin]{footmisc}

\ifcamera \usepackage[accsupp]{axessibility} \fi

% ...

\ifarxiv  \fi

\newcommand{\R}[1]{{%
    \textbf{%
        \ifstrequal{#1}{1}{\textcolor{red}{R#1}}{%
        \ifstrequal{#1}{2}{\textcolor{blue}{R#1}}{%
        \ifstrequal{#1}{3}{\textcolor{magenta}{R#1}}{%
        \ifstrequal{#1}{4}{\textcolor{teal}{R#1}}{%
                           \textcolor{cyan}{R#1}%
        }}}}%
    }%
}}

\usepackage{xr-hyper}

\makeatletter
\newcommand*{\addFileDependency}[1]{
  \typeout{(#1)}
  \@addtofilelist{#1}
  \IfFileExists{#1}{}{\typeout{No file #1.}}
}

\makeatother
\newcommand*{\myexternaldocument}[1]{
    \externaldocument{#1}
    \addFileDependency{#1.tex}
    \addFileDependency{#1.aux}
}

\definecolor{cvprblue}{rgb}{0.21,0.49,0.74}
\usepackage[pagebackref,breaklinks,colorlinks,allcolors=cvprblue]{hyperref}
\usepackage[capitalize]{cleveref}
\crefname{section}{Sec.}{Secs.}
\crefname{table}{Table}{Tables}
\crefname{figure}{Fig.}{Figs.}

\ifarxiv \crefname{appendix}{App.}{Apps.}
\else \crefname{appendix}{Suppl.}{Suppls.} \fi

\usepackage{xcolor}         
\usepackage{colortbl} 
\usepackage[table]{xcolor}

\unless\ifarxiv \myexternaldocument{_supplementary} \fi

\begin{document}
\title{\paperTitle}
\author{\authorBlock}
\twocolumn[{
\maketitle
\begin{center}
\includegraphics[width=0.98\linewidth]{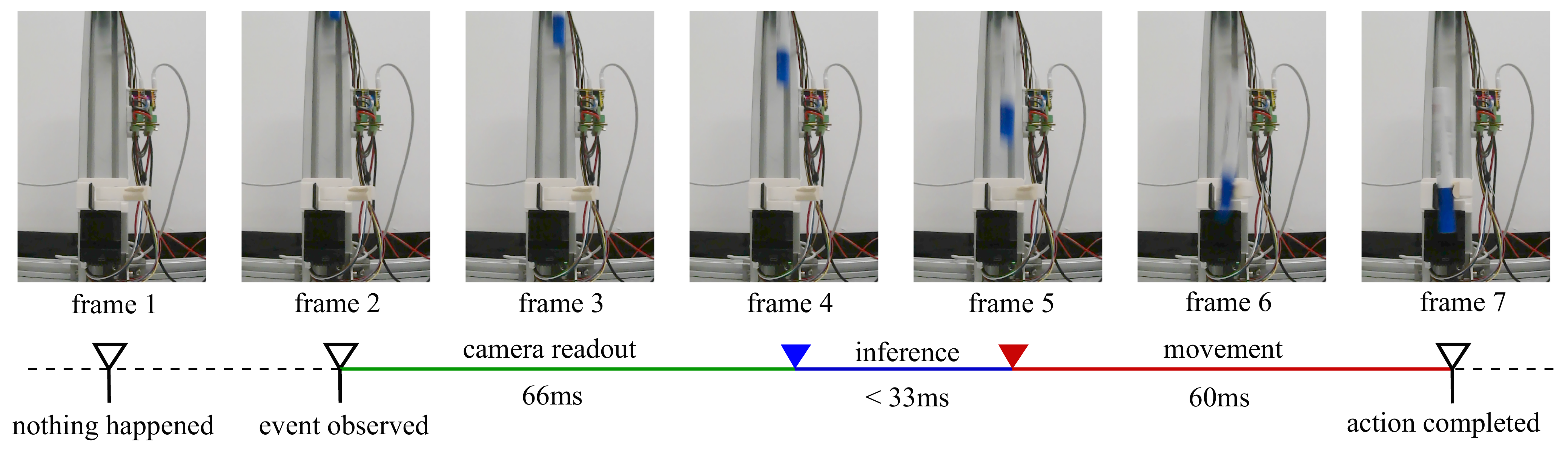} 
\captionof{figure}{Grasping a falling pen. The task has a very stringent time constraint. After observing the pen coming, these is only little time before the action must be initiated. We implemented \textbf{30 FPS inference of VLA model} so that all frames in our camera stream can be processed, and the end-to-end reaction time can be shorter than 200 ms. This is on par with an average human in this test.\label{fig:demo}
\label{fig:teaser}}
\end{center}
}]

\begin{abstract}

In this paper, we show how to run $\pi_0$-level \textbf{multi-view} VLA at \textbf{30Hz} frame rate and at most \textbf{480Hz} trajectory frequency using a single consumer GPU. This enables dynamic and real-time tasks that were previously believed to be unattainable by large VLA models. To achieve it, we introduce a bag of strategies to eliminate the overheads in model inference. The real-world experiment shows that the $\pi_0$ policy with our strategy achieves a \textbf{100\%} success rate in grasping a falling pen task. Based on the results, we further propose a full streaming inference framework for real-time robot control of VLA. Code is available at \url{https://github.com/Dexmal/realtime-vla}. 

\end{abstract}

\section{Introduction}

Learning-based robotic control algorithms are prevailing, especially billion-parameter VLA models~\cite{black2024pi_0,black2024pi_05, shi2025memoryvla, sun2025geovla}. Despite their impressive generalization, these models face a latency issue. Many real-world tasks, such as grasping a moving object, tend to require a quick reaction time. However, a forward pass of VLA models typically requires hundreds of milliseconds, impeding quick reactions that were expected from a dynamic robot.
Running less than 33 ms $\approx 1/30$s is the turning point that enables real-time operation, which means all frames of a 30 FPS RGB video stream can be fully processed. Even if 34 ms is achieved, we must drop frames every now and then during continuous operation. If the event we need to detect happens to occur at the dropped frame, the latency will be increased by a whole frame time.

In this paper, we make a crucial observation that VLAs are indeed capable of running in real-time under a single consumer RTX 4090 GPU. After our optimization, we achieve a latency of 27.3 ms given two input views, significantly faster than ``official'' inference provided by the \texttt{openpi}~\cite{black2024pi_0} project (See Tab.~\ref{tab:speed}).
\begin{table}
\centering
\begin{tabular}{lrrr}
\textbf{Methods} & \textbf{1 view} & \textbf{2 views} & \textbf{3 views}\\
\hline
naive torch & 105.0 ms & 106.5 ms & 113.9 ms\\
openpi/jax & 43.8 ms & 53.7 ms & 67.6 ms\\
\rowcolor{gray!20} ours & 20.0 ms & 27.3 ms & 36.8 ms\\
\end{tabular}
\caption{\label{tab:speed}
$\pi_0$~\cite{black2024pi_0} inference speed comparison on a single RTX 4090. The measurements above assume empty prompt text and 63 chunk length. We push the inference time to camera frequency.
}
\end{table}
The performance gain comes from our engineering of the inference pipeline.
First, we use the CUDA graph approach to eliminate all CPU overheads. Then we make transformations on the computational graph to reduce the total MAC workload or the number of kernel launches.
After that, we rearrange the memory and tensor operations inside individual kernels to better exploit parallelism.
With all of the above strategies, we successfully push inference time to 30 FPS and above, meeting the demand for real-time control.

To verify the effectiveness of our real-time strategy, we design a simple proof-of-concept experiment in real world.  As shown in Fig.~\ref{fig:demo}, two vertically aligned grabbers are made to hold a marker pen.
After the first grabber above releases the pen, the second grabber is required to catch the pen at the right time. Hundreds of grasping data is collected by automatic rules. The $\pi_0$ model is trained to control the grabber to grasp a pen falling from a higher disturbed position. Such a task has a very stringent time constraint. During model inference, the $\pi_0$ model achieves a 100\% success rate on the job thanks to the largely optimized inference time. 

This result encourages us to rethink how to adopt the VLAs into a real-time robot system. Currently, the robotic control system mainly includes three layers and there are different algorithms operating with a hierarchy of control frequencies. The VLAs are believed to live in the layer of mid-level control. Controls of higher frequencies, namely force or torque control, are believed to be handled by other algorithms. However, we find that the VLA itself contains different tiers of input and output frequencies. We directly map the structure of VLA into a full control algorithm and call it the Full Streaming Inference mode. The system is capable of generating control signals at maximum 480 Hz. It touches the threshold of real-time force control.

Extending our framework to do more challenging real-time tasks remains for future researchers. Researches may further introduce the vision-language model (VLM) as the System I on top of VLA part with 1-10 Hz frequency for high-level scene perception and task planning. They may also consider introducing multi-sensory information, such as force and visual-tactile, into the action expert for real-time action chunking decoding.

\begin{figure}
    \centering
    \includegraphics[width=0.95\linewidth]{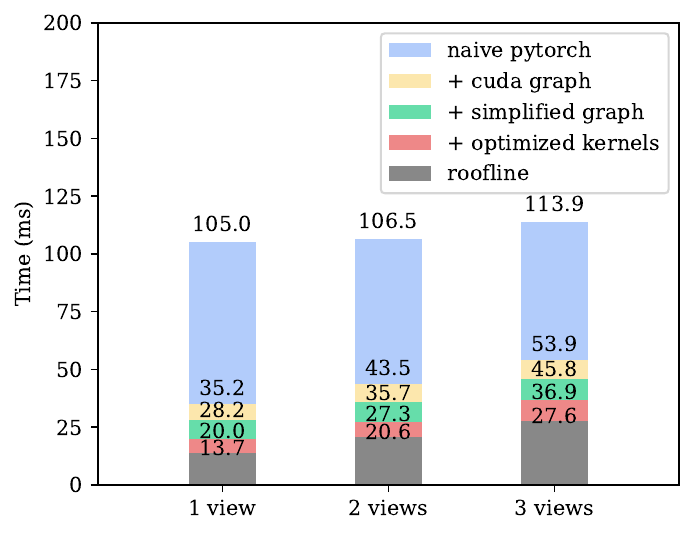} 
    \caption{
    Breakdown of the model running time. From a plain naive pytorch implementation, we show how to reduce redundant computation and eliminate CPU overhead (Sec.~\ref{sec:overhead}). Then we use techniques to optimize the individual kernels (Sec.~\ref{sec:kernels}). Finally we establish a lower bound (Sec.~\ref{sec:lowerbound}) that is not far from the current implementation.
    \label{fig:speed_progress}
    }
\end{figure}

\section{Preliminary on $\pi_0$-level Model}
$\pi_0$~\cite{black2024pi_0} is an excellent vision-language-action (VLA) policy for general robotic manipulation. Hybrid training on both robotic and multimodal data is performed for open-world generalization~\cite{black2024pi_05}. From the perspective of model architecture, it mainly includes two parts: vision-language model (VLM) and action expert (AE). 

\noindent \textbf{VLM} backbone is initialized with PaliGemma~\cite{beyer2024paligemma}, which is a multimodal model with 3B parameters. It is composed of a vision encoder SigLIP~\cite{zhai2023siglip} with 400M parameters and a large language model (LLM) Gemma~\cite{team2024gemma} with 2.6B parameters. The representations of PaliGemma are learned from large-scale web-data pretraining, providing strong prior for parallel action decoding of AE part below.

\noindent \textbf{AE} is coupled with the VLM backbone through the mixture-of-expert (MoE) architecture~\cite{shazeer2017moe}. The multiview images and task prompt are routed to the larger VLM backbone, while the states and action noises are routed to the AE. The network of AE is downsized from Gemma with smaller width and MLP dimension, resulting in 300M parameters. AE is modeled through flow matching~\cite{lipman2022flow} to produce the prediction of action chunking.

\section{Eliminating the Overheads}
\label{sec:overhead}
In the following sections, we show a step-by-step procedure to construct the inference program. Our starting point is a plain pytorch \texttt{nn.Module} implementation that literally follows the model structure. The measured running time is above 100 ms, which is far from our goal.
The first steps we take focus on some ``low hanging fruits'' that drastically speed up computation by getting rid of CPU overhead (the ``+cuda graph'' entry in Fig.~\ref{fig:speed_progress}) and removing redundant computation (the ``+simplified graph'' entry).

\subsection{Removing CPU Overhead}

Nowadays, neural network inference is commonly driven by Python code that launches underlying CUDA kernels. However, the Python part has significant overhead when the number of kernels is large. In $\pi_0$ model, the total number of kernels launched per inference step is estimated to be more than a thousand, which makes the CPU issue urgent.

There are several Ahead-Of-Time (AOT) or Just-In-Time (JIT) compilation techniques available. However, we find the simplest and most effective way is to use the CUDA graph mechanism. In a CUDA graph, we can record the stream of launched kernels during model inference and replay them afterwards. During replay, the kernels are launched purely by GPU and the driver, removing all Python execution overhead. The CUDA graph approach needs to ensure that all kernel codes and buffer pointers are constant from run to run. In our case of VLA, this can be true, as there are no dynamic branches in the underlying transformer blocks.
As shown in Fig.~\ref{fig:speed_progress}, this speeds up inference by around two fold, squeezing out the major part of the inference overhead in our naive implementation.

\begin{figure}
    \centering
    \includegraphics[width=0.95\linewidth]{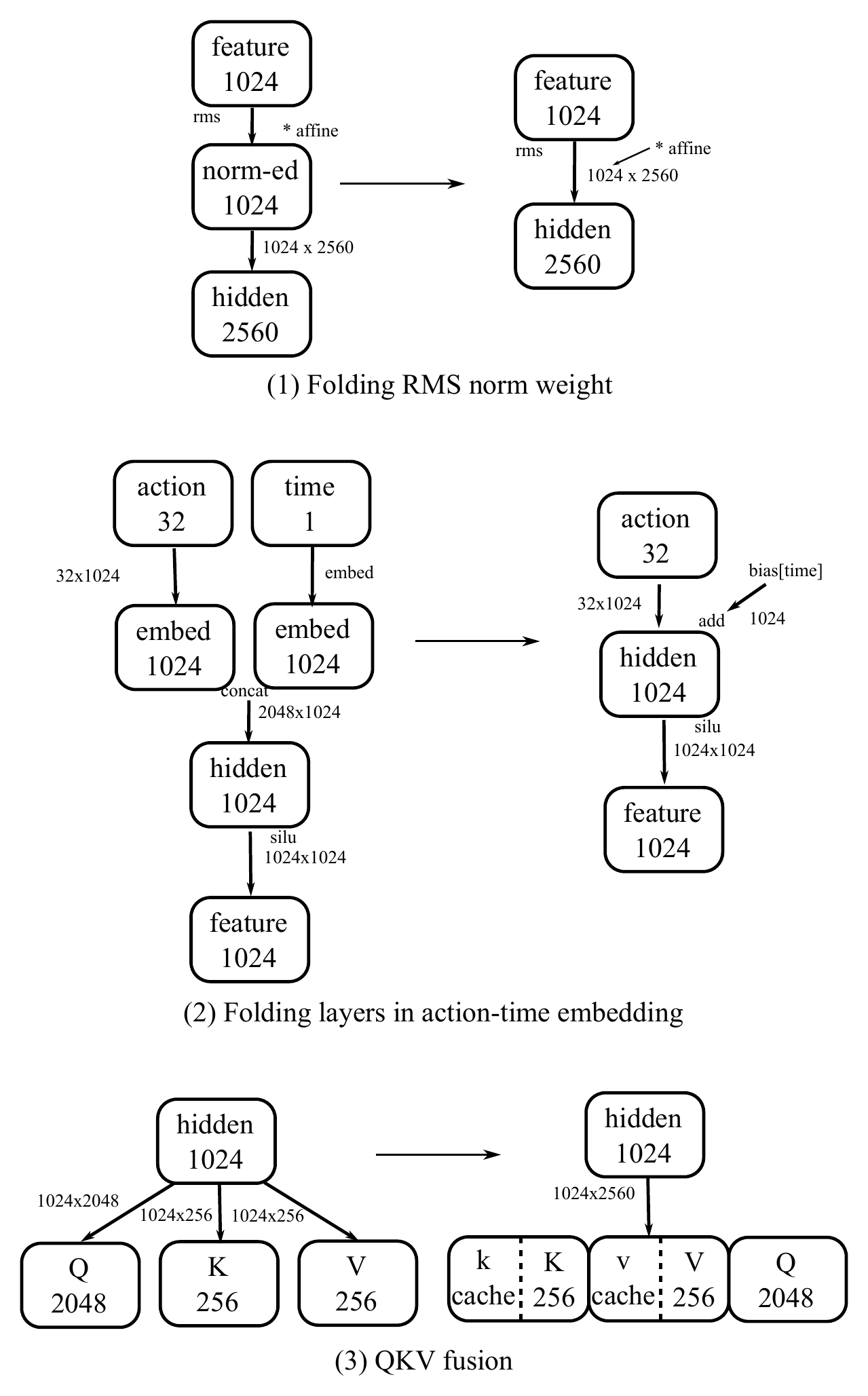}
    \caption{
    Transformations to simplify the computational graph. (1) Absorbing RMS affine parameters to the next linear layer; (2) Folding linear layers in action-time embedding; (3) Fusing QKV as one weight matrix.
    \label{fig:fusiontransform}
    }
\end{figure}

\subsection{Simplifying the Graph}

In the next step, we take a closer look at the network structure and spot some computation that can be rewritten in an equivalent way but runs faster. In compiler literature this is reminiscent to the ``constant folding'' technique, but in the context of model inference there are more that we can do. We list all transformations taken and explain them one by one (see Fig.~\ref{fig:fusiontransform}).

The first is to fuse the affine parameters in the RMS norm layer~\cite{zhang2019rms} into the subsequent linear layer. Because both operations are linear, we can use the associativity law to modify the weights in the linear layer to achieve it.
The second is to fold the ``action time encoder'' in the action expert. The action value is up-projected to 1024 dimensions and concatenated with a projected timestep encoding vector. The result is fed into another linear layer. For the action value branch, we can fold the two consecutive kernels into one as there is no nonlinearity between them. For the time branch, as there are only 10 different time steps during inference, we can tabulate the result of the linear layer and fuse it all the way to the bias vector right before SiLU operation~\cite{elfwing2018silu}. This reduces the number of operators and also saves MACs.
The third is to fuse QKV projections together. We can combine the matrices used in Q, K, and V, to a single large matrix and get the individual results back by slicing the resulting tensor. This reduces the number of kernels and increases parallelization. We can also fuse RoPE~\cite{su2024rope} operation into the matrix multiplication, and precompute the weights used in RoPE.
As shown in Fig.~\ref{fig:speed_progress}, these modifications reduce the inference time by 7-8 ms.

\begin{figure*}[t]
    \centering
    \includegraphics[width=1.0\linewidth]{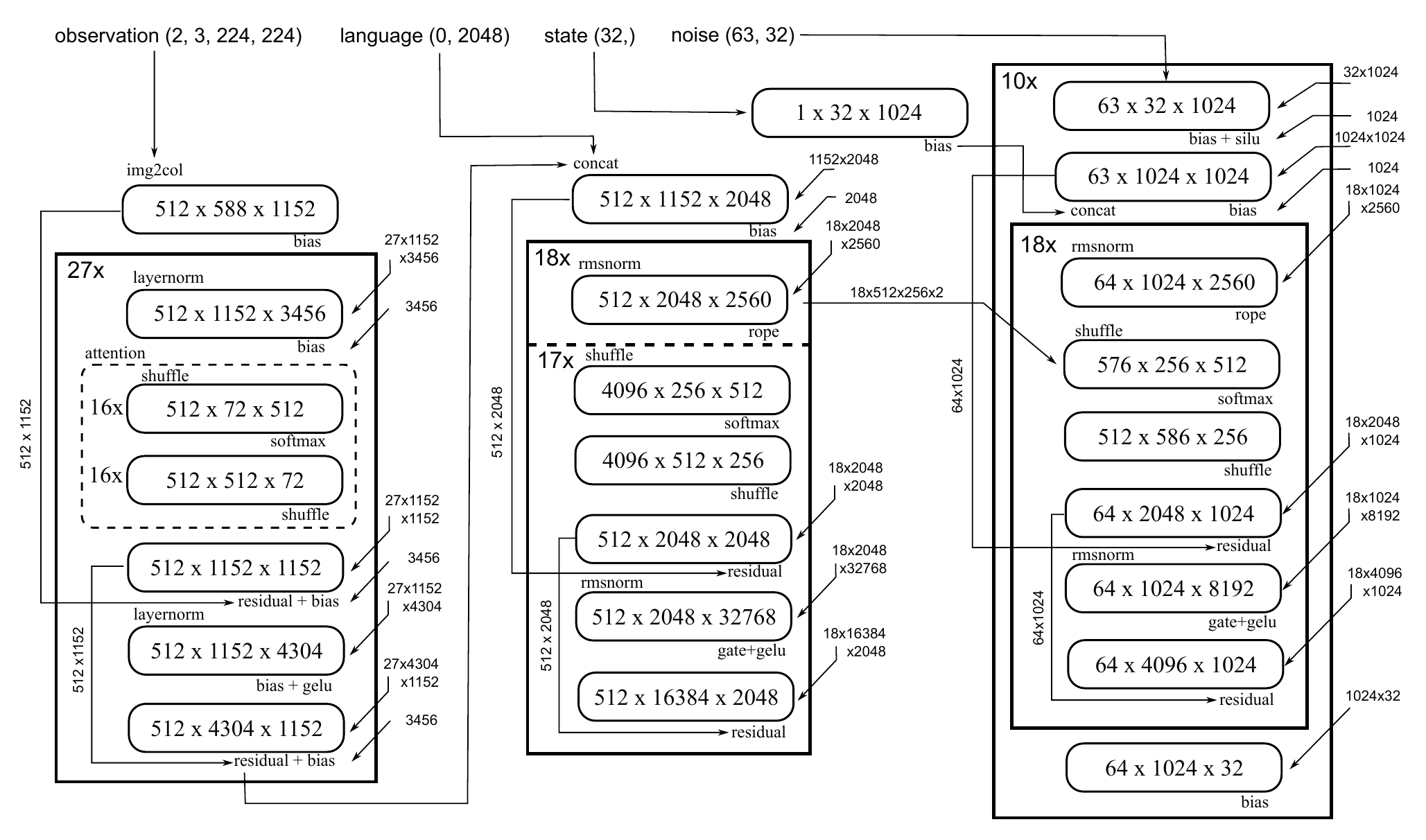} 
    \caption{Computation flow of $\pi_0$ model. The model consists of a vision encoder (left), LLM (middle) and action expert (right). All components can be further decomposed into a series of matmuls and associated scalar operations.}
    \label{fig:matflow}
\end{figure*}

\subsection{Avoiding Other Overheads}

There are other overheads outside the GPU part but are also worth mentioning. The first is image resizing, which can be as slow as several milliseconds if not implemented correctly. For this part, we observe that most camera ISP support many different output resolutions. Using a resolution that is close to the desired 224$\times$224, e.g. 240$\times$320, helps with this issue. Also, carefully writing the resize code by hand reduces inference time significantly, since we find the implementation of JAX version sub-optimal. We measure the time that performing image resizing under correct implementation takes less than 60us on desktop x86 CPU. In the rest of the paper, we do not count the time.

In addition to image resizing, there are other tips for the whole inference system. Copying data back and forth between CPU and GPU requires pinned memory for optimal performance. Making all CPU buffers static can reduce jitter. Using zero-copying to handle camera frames reduces latency. 
This paper do not go further into these details as they are out-of-scope of the paper, but we stress that careful implementation in all parts of the system is crucial for an end-to-end real-time system.

\section{In-Depth Optimization of the Kernels}
\label{sec:kernels}

After harvesting the low-hanging fruits, it is time to attack the ``hard core'' parts. The computation graph after simplification is shown in Fig.~\ref{fig:matflow}. There are 24 GEMM-like operations and associated scalar operators in total. Here are some strategies for further improvements below.

\subsection{Tuning Tile Parameters of GEMM}

The default pytorch implementation of matmul goes to cuBLAS, which dispatches to compiled cutlass kernels according to the matrix dimensions. However, some of the kernels do not receive the optimal configuration. We manually tuned the tiling strategies using a Triton implementation and tabulated the results in Tab.~\ref{tab:kernel_config}, saving about 1.5 ms from Triton optimization.
Note that transformer in LLM only runs 17 times, instead of 18, of attention and FFN layers in the Table. It is because only the KV caches are passed to the AE, so the features of the last layer is not required, which further saves about 0.7 ms.

\subsection{Fusing Gated Linear Layers}

In the transformer part of the model, the FFN uses a gated up-projection implementation. The feature is multiplied by two different weights and the results are combined as $FC_1(x, w_1) \cdot \mathrm{GELU}(FC_2(x, w_2))$. Here, $w_1$ and $w_2$ are the weights for two fully-connected layers $FCs$. During computation, these two matmuls can run parallel, and more importantly, their load and store operations can be coalesced. After loading one input tile, two weight tiles can be loaded for computation. When the results are written back to memory, only the combined result is required to be written, which reduces the memory operation time and improves tensor utilization. In this step, inference performance is further improved by 1.7 ms.

\subsection{Partial Split-k}

In the computation graph (see Fig.~\ref{fig:matflow}), a special GEMM of size $512\times1152\times1152$ is worth mentioning. The core problem of this size is that when using $64\times64$ tile, there will be 144 blocks. It is not a multiple of 128, which means the blocks fail to be distributed to the 128 SMs in RTX 4090. Using smaller blocks helps make the distribution more even, but reduces overall efficiency. Further splitting along the K dimension produces higher overhead.

Our observation is that we can split the GEMM into two parts. The first is a $512\times1152\times1024$ matmul, which can be evenly distributed to the SMs using $64\times64$ tile. The second is $512\times1152\times128$, which can be partitioned to the 128 SMs using $32\times32$ block and split-2 partition in the K dimension. The two parts can be written in a single kernel as they do not depend on each other.
The performance gain from this optimization is less than 0.1 ms, but we think it is an interesting case for further study.

\subsection{Fusing the Scalar Operations}

After optimizing the GEMMs, it is time to deal with the scalar operations. The bias, residual shortcut and activation operations can be trivially combined into the GEMM. For the RMS norm, we first compute the token-level stats into a separate buffer. Then in the next GEMM, we divide the multiplied result by the corresponding factor after all accumulations. These operation reduces the total memory footprint.
The gain in this step is hard to estimate, but we attribute the remaining gain in Fig.~\ref{fig:speed_progress}, which measures roughly 4 ms.

\begin{table*}
\centering
\begin{tabular}{cc>{\columncolor{gray!20}}r|rrl|c>{\columncolor{gray!20}}r}
\textbf{Times} & \textbf{Shape} & \textbf{Roofline} & \textbf{cuBLAS} & \textbf{Triton} & \textbf{Strategy} & \textbf{Oprs} & \textbf{Actual}\\
\hline
1 & 512$\times$588$\times$1152 & 0.004 & 0.036 & 0.044 & triton 64,64,32 & img2col, bias & 0.046\\
27 & 512$\times$1152$\times$3456 & 0.602 & 0.984 & 0.870 & triton 64,64,32 & ln, bias & 0.926\\
27 & attn 16$\times 512^2 \times$72 & 0.178 & / & / & torch & softmax & 0.430\\
27 & 512$\times$1152$\times$1152 & 0.201 & 0.474 & 0.396 & partial split-4 & bias, res & 0.409\\
27 & 512$\times$1152$\times$4304 & 0.750 & 1.221 & 1.074 & triton 64,64,64 & ln, gelu & 1.160\\
27 & 512$\times$4304$\times$1152 & 0.750 & 1.190 & 1.143 & triton 64,64,64,4 & bias, res & 1.158\\
\hline
 & \textbf{Vision Encoder} & \textbf{2.485} & 4.334 & 3.957 &  &  & \textbf{4.059}\\
\hline
1 & 512$\times$1152$\times$2048 & 0.013 & 0.041 & 0.041 & triton 64,64,64 & ln, bias & 0.042\\
18 & 512$\times$2048$\times$2560 & 0.529 & 0.823 & 0.761 & triton 64,64,64 & rms, rope & 0.862\\
17 & attn 8$\times 512^2 \times$256 & 0.200 & / & / & torch & softmax & 0.406\\
17 & 512$\times$2048$\times$2048 & 0.399 & 0.495 & 0.511 & triton 128,64,64 & res & 0.524\\
17 & 512$\times$2048$\times$32768 & 6.391 & 7.359 & 7.317 & fused gate & ln, gate & 7.274\\
17 & 512$\times$16384$\times$2048 & 3.195 & 3.751 & 3.696 & triton 128,64,64 & res & 3.740\\
\hline
 & \textbf{LLM} & \textbf{10.727} & 12.875 & 12.732 &  &  & \textbf{12.503}\\
\hline
1 & 1$\times$32$\times$1024 & 0.000 & 0.027 & 0.026 & triton 16, 16, 32 & bias & 0.026\\
10 & 63$\times$32$\times$1024 & 0.001 & 0.057 & 0.036 & triton 32, 32, 32 & bias, silu & 0.038\\
10 & 63$\times$1024$\times$1024 & 0.021 & 0.072 & 0.060 & triton 32, 32, 64 & bias & 0.063\\
180 & 64$\times$1024$\times$2560 & 0.934 & 1.718 & 1.479 & triton 64, 32, 64 & rms, rope & 1.738\\
180 & 512$\times$256$\times$576 & 0.149 & 0.723 & 0.602 & triton 32, 32, 64 & softmax & 1.071\\
180 & 512$\times$576$\times$256 & 0.149 & 0.883 & 0.554 & triton 32, 32, 64 &  & 0.590\\
180 & 64$\times$2048$\times$1024 & 0.747 & 1.164 & 1.203 & triton 32, 32, 128 & res & 1.237\\
180 & 64$\times$1024$\times$8192 & 2.990 & 3.703 & 3.559 & fused gate & rms, gate & 3.847\\
180 & 64$\times$4096$\times$1024 & 1.495 & 2.367 & 2.226 & triton 16, 32, 256 & res & 2.290\\
10 & 63$\times$1024$\times$32 & 0.001 & 0.055 & 0.048 & triton 16, 16, 256 & res & 0.061\\
\hline
 & \textbf{Action Expert} & \textbf{6.486} & 10.808 & 9.831 &  &  & \textbf{11.001}\\
\hline
 & \textbf{In total} & \textbf{19.698} & 28.017 & 26.520 &  &  & \textbf{27.299}\\
\hline

\end{tabular} 
\caption{Detailed configuration of the kernels. This table assumes two views and no prompt. We list the matmul dimensions and the scalar operations associated with each matmul. We also measured the cuBLAS (accessed by torch.matmul) time of the corresponding matmul, the tile-based triton implementation time, and the actual running time after using the optimal implementation strategy and fusing the scalar operations. The ``torch'' strategy means using plain torch implementation is good enough. The ``triton n,m,k(,split)'' stands for the tile size of the matmul and the potential split-k dimension. There are also other special strategies that are described in Sec~\ref{sec:kernels}.\label{tab:kernel_config}}

\end{table*}

\section{Establishing Lower Bound}
\label{sec:lowerbound}

In this section, we show how far we are from an ``ideal'' implementation. Our arguments are based on two parts: the unavoidable time for the matrices, and the non-negligible time of synchronization.

\subsection{Roofline of the GEMMs}

The basic approach is the ``roofline'' model, which lower bounds the time of a computation by the maximum of doing all memory operations and all compute operations. In the context of model inference, this translates to calculation of the HBM bandwidth and tensor core cycles.

For one BF16 GEMM operation of dimension $N\times K \times M$, the lower-bound is as follows:
$$
t_{\mathrm{roofline}} = \mathrm{max}(\frac{2K  M}{T_\mathrm{bandwidth}}, \frac{N K  M}{T_\mathrm{compute}}).
$$
In the above equation, we consider memory operation only for the second matrix. It is because in a network the first matrix and the resultant matrix are usually the activation features, which can be allocated on the L2 cache. The network parameters on the other hand, are too large to be fully cached so we should account for it.

For RTX 4090, the claimed memory bandwidth is 1.01 TB/s. The claimed BF16 MAC/s (with FP32 accumulation) is 82.6 T, but we observe that the using card has a boosted frequency of 2.79G Hz and hence 91.4 TMAC/s in total. These parameters are used in our computation in Tab.~\ref{tab:kernel_config} (for 2 input views).
Summing the roofline values gives the following results:

\begin{center}
\begin{tabular}{c|c|c}
  \textbf{1 view}   & \textbf{2 views} & \textbf{3 views} \\
\hline
  12.8 ms & 19.7 ms &  26.7 ms
\end{tabular}
\end{center}

We need to note that it is tricky to sum two consecutive operations when using the roofline model. Theoretically, for two sequential matmuls (e.g. in FFN), there is indeed chance to start part of the computation of the second matmul before the first finishes. Then we need to take the two matmuls as a whole to apply the roofline model, which mathematically may be lower than the sum of the roofline values of the individual operators. However, we see a clear trend that most operations in vision encoder and LLM are computation-limited, and most operations in the AE are bandwidth-limited. Therefore, overlapping work across consecutive operators does not affect the roofline argument for most cases. Thus, it is proper to use the sum-of-parts approach.

\subsection{Synchronization Overhead}

We notice that there are a large number of kernels launched in the CUDA stream. For data consistency, the SMs need to wait for others to complete one kernel before launching the next, which produces synchronization overhead.

There are totally 1378 matmuls show in Fig.~\ref{fig:matflow} and an experiment is performed to estimate the synchronization overhead. We run a simple kernel for 1378 times and we write one kernel that runs the computation inside a for loop for the same amount of time without any synchronization, as comparison baseline. The additional time after considering synchronization should be an estimate of the overhead in model inference.
We tested different synchronization methods. The first is consecutive kernel launches in the naive pytorch code. It also contains CPU overhead, so the value is large. We used the CUDA graph to chain all kernels and get a result of 1.72 ms, which is close to our experience. The third approach is a ``software barrier'' approach that we will explain below.

\begin{table}
\centering
\begin{tabular}{crr}
    \textbf{Setting} & \textbf{Time} & \textbf{Overhead} \\
    \hline
    Pytorch & 13.81 ms & +12.92 ms \\
    Cuda graph & 2.61 ms & +1.72 ms \\
    Software barrier & 1.75 ms & +0.86 ms \\
    Fused no sync & 0.89 ms & +0 ms \\
\end{tabular}
\caption{Measuring overhead time between kernels. We run a simple A+B computation on $256\times 1024$ elements using 256 blocks for 1378 times. We use a fused Triton kernel without any synchronization as the baseline time. Then we implement the computation with different methods, and measure $t_\mathrm{sync}$ as the extra time compared to the baseline.\label{tab:overhead}}
\end{table}

The traditional view of the CUDA programming model says we cannot explicitly synchronize across all blocks, as they may not live together at all. However, in practice if we launch a number of blocks that equals the number of SMs, we can make sure that all blocks run together and use global memory to create a barrier. An example code (in Triton) is as follows:

\begin{verbatim}    
lock_goal += psize
tl.atomic_add(lock_ptr, 1)
while tl.atomic_or(lock_ptr, 0)\
      < lock_goal:
    pass
\end{verbatim}

Then to synchronize between the two kernels, we can write the two kernels as one and insert the barriew in between. We find the approach above to be faster than the CUDA graph approach. The overhead becomes 0.86 ms. However, using the software barrier has a few consequences. First, the grid size of the fused kernels need to be matched. Second, the number of required registers and shared memory may increase. Third, the code size of the kernel is increased, incurring overheads in other parts of the system. In practice, we tried to fuse operators using the software barrier but find negative performance gain. Hence we only use it as the indicator of synchronization cost in the ``lower bound'' argument.
After counting for the additional 0.86 ms, our updated lower bounds become:

\begin{center}
\begin{tabular}{c|c|c}
  \textbf{1 view}   & \textbf{2 views} & \textbf{3 views} \\
\hline
  13.7 ms & 20.6 ms &  27.6 ms
\end{tabular}
\end{center}

Compared to our performance in Fig.~\ref{fig:speed_progress}, we see the remaining room of improvement is at most 30\%. It means that we have been close to an optimal implementation.

\section{Full Streaming Inference}
\begin{figure*}[t]
\centering
\includegraphics[width=0.9\linewidth]{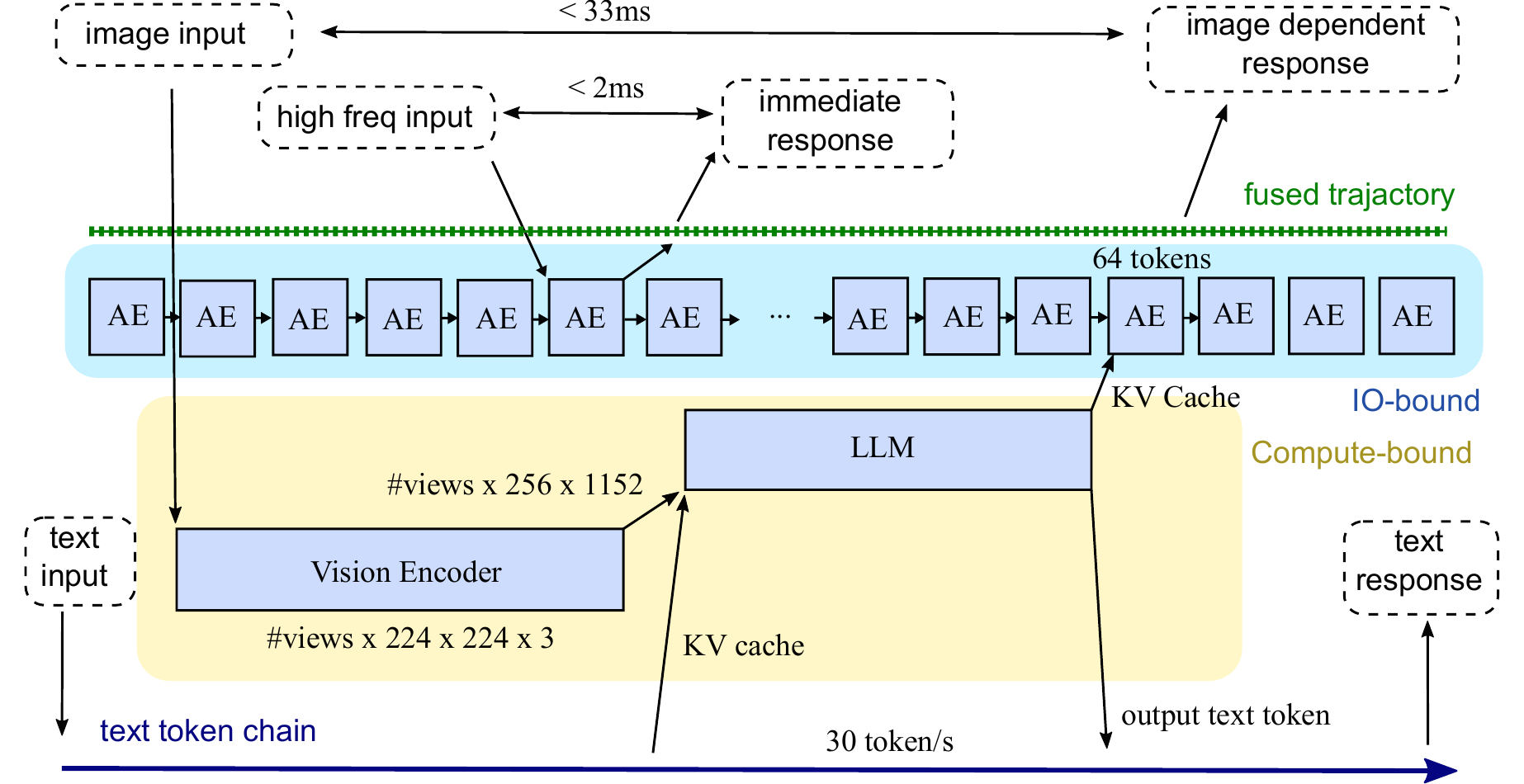}
\caption{The Full Streaming Inference framework. AE denotes the action expert. 
\label{fig:framework}}
\end{figure*}

In this section, we will go further to explore the possibility of other control frequency loops beyond 30 FPS.

\subsection{Overlapped Streaming Inference}

Our core observation is that \textbf{overlapping kernels increases throughput}. Specifically, we consider the AE kernel and VLM kernel. The former is IO bounded, and the latter compute bounded. If we run the kernels together, both IO and compute resource would be better utilized. This means from a computational perspective, finding a way to run them concurrently, or, \textbf{streaming} them, would be beneficial.
We do a proof-of-concept measurement in the following table:

\begin{center}
\begin{tabular}{l|l}
\textbf{Condition} & \textbf{Time} \\
\hline
Sequential VLM + 10 AE & 27.3 ms \\
Concurrent VLM + 10 AE & 26.3 ms \\
Concurrent VLM + 16 AE & 32.7 ms \\
\hline
\end{tabular}
\end{center}

In this measurement, we assume two input views, and the baseline is running the 10 action-expert steps after VLM. If we create two CUDA streams and push the AE kernels in parallel to the VLM using older KV cache, the total running time decreases (the second line). We can increase the number of AE kernels up to 16 runs until the total running time touches the 1/30s bound (the third line).

This reveals a fact: so long as we can run the kernels concurrently, we can afford \textit{30 VLMs and 480 AEs} in one second. Moreover, the AEs are running right next to each other, spanning evenly on the timeline. This opens up a possibility of creating an additional control loop at 480 Hz.
However, realizing this idea requires far more than writing kernels. We need to totally restructure the role of VLM and AE in a VLA. We term this line-of-thought the Full Streaming Inference paradigm, as illustrated in Fig.~\ref{fig:framework}. In the next few subsections we will go through the details.

\subsection{Re-thinking Action Expert: Up to 480 Hz}

It should be noted that the \textit{frequency of the output trajectory} and \textit{frequency of control} are two totally different concepts. The first one means the density of the nodes in the output trajectory, and can be trivially increased by interpolation. The latter demands a bounded time from stimuli to reaction, which requires high-frequency processing along all nodes in the processing pipeline.

To achieve 480 Hz control frequency, we first need to find an observation signal that can be sampled at that frequency. Luckily, nowadays force sensors, being 3D or 6D, are all capable of generating samples at over 2K Hz, and the latency inside the sensor is extremely low (down to us level). So this kind of signal is our best candidate for injection into the network. For a system without force sensors, we can consider motor current (which is typically controlled at beyond 1K Hz), or resistance-based tactile signals.

As seen in Fig.~\ref{fig:framework}, the computational paradigm tells us that we need to \textbf{inject the high-frequency input signals into the AE}. This means, there will be a signal that is ``seen'' by the AE, but is not observed (or at least not observed immediately) by the VLM. We would expect the AE be able to immediately generate an action (e.g. emergency stop) when there is a clear sign in the input signal.
This will introduce a shift in the role of AE. The ``traditional'' usage of the AE is to implement the flow matching or diffusion integration. The output is not ready for use until all the denoising steps are completed. The AE generates a full list of actions only after the 10 denoising steps. However, it is not hard to see 
that we can rewrite the flow matching algorithm so that the generation is ``gradual'' -- each step generates a part of the action list, as if we are doing auto-regressive decoding. The Real-time chunking (RTC)~\cite{black2025rtc} algorithm has already been using similar techniques. 

After doing this change to the AE, the remaining parts would be straightforward. We can implement the injection of new input signals as addition of a new input token to the transformer-based AE. Whenever a new sample comes from the sensor, we can do a memcpy operation in a separate stream to update the corresponding value in the GPU global memory. The execution of the VLA can be totally transparent of this update.
On the output side, we can view the AE as continuously manipulating consecutive timestamped nodes in a dense 480 Hz trajectory. The nodes will be ``committed'' when they are retrieved out of the GPU, in a potentially asynchronous fashion. The AE will only change future uncommitted action values according to the flow rules.


After the discussions above, we can go back to Fig.~\ref{fig:framework}. We have discussed three main objects in the picture. There is a \textbf{trajectory buffer} of 480 nodes per second that can be updated and retrieved. A \textbf{stream of action experts} run at 480 Hz, receiving input from (1) the former AE run (2) the high frequency sensors (3) the KV cache from the most recent VLM run, and processes up to 64 tokens. This tokens are fused back to the trajectory buffer according to some update rule, which means the ``window'' of a AE can cover 4 frames if the tokens are one-to-one with the nodes. We do not discuss the detail of the fusion in this report. The \textbf{stream of VLM} run at 30 Hz, converting latest image frames into KV cache for the AE to use.

There are two feedback loops in the picture. The \textbf{quick loop} is the injection of the high frequency signal, processing of one AE, and generation of the reaction trajectory. This can be as quick as 2 ms in the most favorable case. The \textbf{``slow'' loop} is the image-driven loop. After a frame is captured, it must be processed by the VLM. When the result is used by at least one AE, we will see its influence to the robot action. Thus the feedback time can be as low as 1/30 s in the most optimistic case.
The two loops are run in parallel. Their overlapped execution improves utilization of the hardware. And as they all run continuously (and potentially fully asynchronously), we will call this running method a Fully Streaming one. The best way to implement this may be the Persistent Magakernel approach, and due to the scope of this report we will not go into further detail.

\subsection{Fusing VLMs: Going below 1Hz}

Modern VLAs not only process visual data, they also receive and generate text, either in the form of multi-modal understanding, task planning and CoT reasoning.
We make one further observation to show how this can be incorporated into our paradigm. We have already been ``encoding'' the visual tokens at 30 Hz. This means 30 passes over the transformer weights. The text inference in LLM, the bottleneck is the loading of model checkpoint during autoregressive inference. So we can \textbf{piggyback} the text inference with frame encoding. After loading one matrix weight, it is first used to compute the matmul in the VLM part of VLA, and then compute the inference of the text data. This can be implemented as easy as a special attention matrix. Since the number of visual tokens is large, the additional MACs should not alter the computation time much.

The network result of the above modification is that we have one additional auto-regressive text stream of 30 token/s, as shown in the bottom row in Fig.~\ref{fig:framework}. We can use this quota for interaction with the user, or letting the model do reasoning. 30 tokens/s is a fair amount, as humans can only speak at about 3.3 tokens/s.

\subsection{Summary}

Summing up the discussions above, there is a chance to create three feedback loops in a real-time VLA system:
\begin{itemize}
    \item \textbf{The force loop at 480 Hz.} High-frequency input and output are handled by AE inference.
    \item \textbf{The visual loop at 30 Hz.} Image-based reaction handled by the VLM.
    \item \textbf{The textual loop below 1 Hz.} Text-based interaction and reasoning at a lower frequency, bringing more intelligence.
\end{itemize}

\noindent In this report, we only give a sketch of the design of this framework. All implementation details are omitted, and this will be the content of subsequent works.

\section{Real World Validation}

We design a simple real-world experiment to verify the effectiveness of our implementation. 

\subsection{Setup and Data Collection}

The hardware setup is shown in Fig.~\ref{fig:teaser}. Two custom grabbers, which are aligned vertically, are made to hold a marker pen. After the first grabber releases the pen, the second grabber will catch the pen if it closes at the right time. Closing too soon or too late will leave the pen outside. We set the grabber to close or open within 60 ms, which roughly corresponds to two camera frames.

A 30 FPS 720P USB camera is installed to observe the apparatus. The camera sees the second grabber and the pen, but it will not see the first grabber. So it has no direct observation of the time the first grabber releases the pen. We measured the delay of the camera is about two frames. We surmise that one frame is delayed in ISP processing, and the second in USB transmission. Certainly we have access to cameras of higher framerates or lower latency, but for the sake of demonstration we will stick to this choice of the camera. It should be noted that the RealSense cameras commonly used in robotic research are measured to have a delay of over 100 ms, so we decide not to use them in this experiment.

The pen travels about 30 cm before it is caught. We let the first grabber hold different positions of the pen to create a small variation in the falling distance. When we record demonstration data, the second grabber always closes 200 ms after the first grabber releases. We observe that this strategy success 100\% of the times in our demonstration data recording runs. The pose of the camera is slightly adjusted during different runs to create variation.

As the primary recorded data is the expected time to close the grabber for each frame, we post-process them to 1D trajectories describing the target grabber state after the time of the frame. A value of 0 means the grabber should take no action at that time, and a value of 1 means the grabber should close.
We totally collected 600 episodes for training. The camera frames are zero-padded to rectangular shapes to fit network input. Because we allow two input views in our system, we feed the network with both the current frame and the previous camera frame. This also gives the network a hint of the current speed of the pen.

\subsection{Training and Inference}

We use the official \texttt{openpi} repository\footnote{\url{https://github.com/Physical-Intelligence/openpi}} to train our model. The prompt is set to empty. Each training episode includes a few seconds before the pen is released and a few seconds after the pen is caught. Due to adequate number of episodes, we only train the network for a few epochs.

Our inference program consists of different threads to handle input and output. A camera thread waits for the frame to come, and put the frame data (and the time stamp of the arrival of the frame) into a ring buffer. The inference thread always take the latest frame data and runs the network. The output is a trajectory of timestamped grabber states. We also use a circular buffer for storing the output states at different time slots. The inference thread always overwrites the old values in the output buffer. The last thread loops forever to send the corresponding item in the output buffer to the grabber. We carefully implement the code to eliminate any unnecessary delay in the observation-to-action path.
We found the GPU card needs warm-up to reach full speed (power consumption and clock rate). So in our experiment we always start the release after the inference time stabilizes. 

\subsection{Result}

We expect the network to learn from the image the remaining amount of time the pen reaches the grabber. For modern VLAs, this task is so simple that it learns quite well. We perform 10 consecutive experiments to test the system, and all catches are successful (100\% success rate).

From a learning perspective, this result is trivial. We would expect the same result had we used LeNet~\cite{lecun2002lenet}, or even SVMs~\cite{hearst1998svm}. However, from a system point of view, a single successful catch verifies the low latency of our VLA implementation. Due to the capacity of the model, we would expect the system capable of handling more challenging tasks. It is worth noting that we also tried to let humans grab a pen falling from other's hand. We observe that 30 cm is a common lower bound on the travel distance of the pen for a human to react.

In conclusion, we have demonstrated a system that reacts as quickly as human, and at the same time contains billions of parameters in the network. We believe this result should open up the possibility of studying the VLAs on various time-critical tasks.

\section{Related Work}
\textbf{Manually-designed Kernels:} Low-level GPU acceleration has traditionally relied on gpu experts to manually designed kernels provided by vendor-optimized libraries. Frameworks such as XLA~\cite{sabne2020xla}, TensorRT~\cite{developernvidia}, and NVIDIA CUTLASS~\cite{2023cutluss} improve performance through operator scheduling, kernel autotuning, and highly optimized GEMM primitives. Some tensor complilers,like Triton~\cite{tillet2019triton}, decompose tensor into tiles to map directly to the hardware. Those frameworks leave scheduling and operator fusion to the programmer so consequently leave significant performance penalties from kernel-launch overhead and repeated global-memory movements. For transformer-based models, attention computation is widely accelerated by fusing and sheduled particularly. FlashAttention~\cite{dao2022flashattention} uses tiling to reduce the number of memory I/O between HBM and SRAM and fuses softmax with matmul through a repair computation. FlashDecoding~\cite{dao2023flash} proposes a long-context inference strategy that partitions attention into two reduction kernels, reducing KV memory movement and improving auto-regressive decoding efficiency. For AE acceleration, those methods overlook the other parts like FFN. These approaches still require multiple kernel launches and frequent DRAM accesses across the rest of the model.

\noindent \textbf{Scheduling-based approaches:} Schedule-based kernel compilers such as TVM~\cite{chen2018tvm} decouple the algorithm from its execution schedule, enabling loop tiling, fusion, and layout transformations to be applied independently for GPU optimization. These systems leverage user-defined schedules or automated search to improve performance across diverse operators. However, they still require explicit algorithm specifications for each kernel, and their effectiveness is fundamentally constrained by the quality of these provided algorithms and schedules. 

\noindent \textbf{Superoptimization Approaches:} Recent work has explored applying superoptimization to tensor programs, enabling transformations beyond traditional schedule tuning. Among them, Mirage~\cite{wu2025mirage} and Neptune~\cite{zhao2025neptune} perform hierarchical optimization across thread, tile, and operator levels, combining algebraic and schedule transformations to search for high-performance kernels. While this multi-level joint optimization allows them to outperform existing kernel-level superoptimizers, mirage relies on statistical equivalence checking for correctness and they often requires long search times or low acceleration ratio, limiting their practicality for real-time VLA inference.

\section{Future Directions}
Recently, RTX5090 has been released, showing more computation power than ever. Also, there are other edge-oriented accelerators joining the race. This raises the question that if we have more compute at hand, what other outcomes should we expect from the system?
Here we list some possibilities and briefly describe the key issues towards realizing them.

\subsection{Visual-based Latency: Going to 60-120 FPS}

The first direction is to increase the camera FPS and number of views. This corresponds to faster encoding of the network. As the encoding stage is already compute-bound, all we can hope is a sheer increase in the MACs.
One computation alternative is to use lower precision. All results in this report use BF16. If 8-bit multiplications work, the compute power will be unleashed by a significant amount. Quantizing VLAs to low bit precision will be a good research topic.

Another direction is adaptive selection of views. A typical bimanual robot has at least three cameras. In theory, we can dynamically fuse their information into fewer number of tokens if we take into account the current ``active'' view. This requires careful engineering of the training and inference pipeline. However, if it is done right, we would expect great boost in FPS.
Humans are reported to be able to distinguish between 30 FPS and 60 FPS video playbacks. This suggests that 60 FPS may be the next stage in our pursuit. Going to 120 FPS will ensure that our system is ``beyond human'', as this format is usually used for Slow Motion in video recordings.

\subsection{Larger Models: Going to 7B and Beyond}

Another possibility is to scale the model size. Again, scaling VLM means increase in total MACs, and scaling AEs demands increased bandwidth.
RTX 5090 features much more bandwidth (1.79TB/s) improvement than BF16 MACs compared to its predecessor. So putting more parameters in the AE should not be too much a problem, though it is unclear how scaling AEs benefit model learning.

We think 7B parameters should be a good next mile-stone. Its increase against the current 3B $\pi_0$ is not too far that we can still think of techniques to make it run, yet the model performance of the 2x increase in parameters have been well studied in LLM literature.

\subsection{More Fine-grained Feed-back Loop}

We note that in our current construction, there is still a loop running at higher frequency than the AE. It is the layers in the AEs. There are multiple of thousands of layers running in one second. If we have a way to let each layer be aware of the most recent signal, and finding a way for quicker intermediate output, there may be a chance of even higher control frequencies. However, we currently have not seen a way to do this. Also, collecting demonstration data of kilos of samples in one second is a challenge on its own right.
We leave exploration of higher frequencies in the framework of VLA to future researchers.

{\small
\bibliographystyle{ieeenat_fullname}
\bibliography{11_references}
}


\end{document}